\documentclass[11pt]{amsart}
\usepackage{amsmath}
\usepackage{latexsym}
\usepackage{amsfonts}
\usepackage[normalem]{ulem}
\usepackage{soul}
\usepackage{array}
\usepackage{amssymb}
\usepackage{extarrows}
\usepackage{graphicx}
\usepackage{algorithm}
\usepackage{algorithmic}
\usepackage{adjustbox}

\usepackage{wrapfig}
\usepackage{subfigure}
\usepackage{wrapfig}
\usepackage{wasysym}
\usepackage{enumitem}
\usepackage{adjustbox}
\usepackage{ragged2e}
\usepackage[svgnames,table]{xcolor}
\usepackage{tikz}
\usepackage{longtable}
\usepackage{changepage}
\usepackage{setspace}
\usepackage{hhline}
\usepackage{multicol}
\usepackage{tabto}
\usepackage{float}
\usepackage{multirow}
\usepackage{makecell}
\usepackage{fancyhdr}
\usepackage[toc,page]{appendix}
\usepackage{caption}
\usepackage{graphicx}
\usepackage{amsmath}
\usepackage{changepage}
\usepackage{lipsum}
\usepackage[most]{tcolorbox}
\usepackage{multicol}
\usepackage[hidelinks]{hyperref}

\tcbset{
    myboxstyle/.style={
        colframe=gray!50,
        colback=gray!10,
        coltitle=black,
        coltext=black,
        fonttitle=\bfseries,
        boxrule=0.8mm,
        width=\columnwidth,
        sharp corners=all,
        boxsep=4pt,
        arc=2mm
    }
}

\usetikzlibrary{shapes.symbols,shapes.geometric,shadows,arrows.meta}
\tikzset{>={Latex[width=1.5mm,length=2mm]}}

\renewcommand{\_}{\kern-1.5pt\textunderscore\kern-1.5pt}
\title[HardML]{HardML: A Benchmark For Evaluating Data Science And Machine Learning knowledge and reasoning in AI}
\author{Tidor-Vlad Pricope \\ \textit{Independent Machine Learning Engineer}}
\address{Canary Wharf, London, United Kingdom}
\email{ptidor1@gmail.com}
\date{22 January 2025} % submission date 



\begin{document}

\begin{abstract}
We present HardML, a benchmark designed to evaluate the knowledge and reasoning abilities in the fields of data science and machine learning. HardML comprises a diverse set of 100 challenging multiple-choice questions, handcrafted over a period of 6 months, covering the most popular and modern branches of data science and machine learning. These questions are challenging even for a typical Senior Machine Learning Engineer to answer correctly. To minimize the risk of data contamination, HardML uses mostly original content devised by the author. Current state-of-the-art AI models achieve a 30\% error rate on this benchmark, which is about 3 times larger than the one achieved on the equivalent, well-known MMLU-ML. While HardML is limited in scope and not aiming to push the frontier—primarily due to its multiple-choice nature—it serves as a rigorous and modern testbed to quantify and track the progress of top AI. While plenty benchmarks and experimentation in LLM evaluation exist in other STEM fields like mathematics, physics and chemistry, the sub-fields of data science and machine learning remain fairly underexplored.
\end{abstract}

\maketitle
\section{Introduction}

\label{sec:intro}
Recent advancements in large language models (LLMs) have led to significant progress in natural language processing tasks such as translation, summarization, question answering, and code generation \cite{1,2}. These models have been extensively evaluated using benchmarks covering a wide range of subjects, providing valuable insights into their capabilities \cite{3,4}. For instance, the Massive Multitask Language Understanding (MMLU) benchmark \cite{5} assesses LLMs across diverse disciplines, including STEM fields like mathematics, physics, and chemistry \cite{6,7,8}. However, data science (DS) and machine learning (ML) have received relatively little attention in benchmarking efforts. The MMLU test set contains only 112 machine learning questions. Moreover, in the few instances where these domains have been explored, state-of-the-art AI models achieve near-saturation performance, rendering existing benchmarks less effective for distinguishing model capabilities.

It is imperative to devise novel benchmarks that keep up with the rapid advancements in LLMs. This necessity is exemplified in the FrontierMath benchmark \cite{15}, which introduces a future-proof evaluation for mathematics by presenting problems that remain unsolved by over 98\% of current AI models. Such benchmarks are crucial for continuing to challenge and develop advanced AI systems.

Data science and machine learning are foundational to modern artificial intelligence, driving advancements in everything from healthcare to finance \cite{9,10}. Mastery in these fields requires not only theoretical understanding but also practical skills in applying algorithms, statistical methods, and computational techniques to solve complex, real-world problems \cite{11,12}. As AI systems become increasingly involved in DS and ML tasks—ranging from automated model training to data analysis—it is crucial to assess their proficiency and reasoning abilities in these areas. However, as of January 2025, benchmarks in this domain are very limited. The most notable examples include the test ML subsection of MMLU (MMLU-ML) \cite{5}, which consists of 112 multiple-choice questions, and MLE-bench \cite{16}, introduced by OpenAI, which evaluates practical ML engineering skills using a collection of 75 coding questions modeled after Kaggle competitions.

To address this gap, we propose HardML, a benchmark specifically designed to evaluate the knowledge and reasoning capabilities of AI models in data science and machine learning.  HardML employs the same testing framework as MMLU, comprising multiple-choice questions, with the primary difference being that more than one answer can be correct. It differs in scope from MLE-bench, as it does not test coding capabilities but focuses on theoretical understanding and reasoning skills based on theoretical concepts in DS and ML. HardML uses 100 challenging multiple-choice questions, meticulously crafted over six months to cover the most relevant and contemporary topics in DS and ML. The questions are designed to be difficult even for experienced professionals, such as senior machine learning engineers, thereby ensuring that the benchmark assesses advanced understanding and critical problem-solving skills.

A key aspect of HardML is its emphasis on originality and contemporary relevance, featuring questions that reflect the latest advancements in machine learning from the past two years. To minimize the risk of data contamination—where models might have been trained on benchmark content, leading to artificially inflated performance \cite{13}—we developed primarily original questions. By "original," we mean that while the core concepts required to solve these questions may be known (similar to foundational theorems in mathematics), the specific applications and reasoning required are unique. These questions span topics including natural language processing, computer vision, statistics and statistical modeling, classical machine learning algorithms, and more. In this paper, we also introduce EasyML, a benchmark of 85 multiple-choice questions designed to provide a more accessible and slightly easier set of questions than MMLU-ML for evaluating smaller language models, such as GPT4o-mini \cite{21} and LLaMA models with fewer than 70 billion parameters \cite{22}.

\begin{figure}[h]
\centering
\includegraphics[width=1\textwidth]{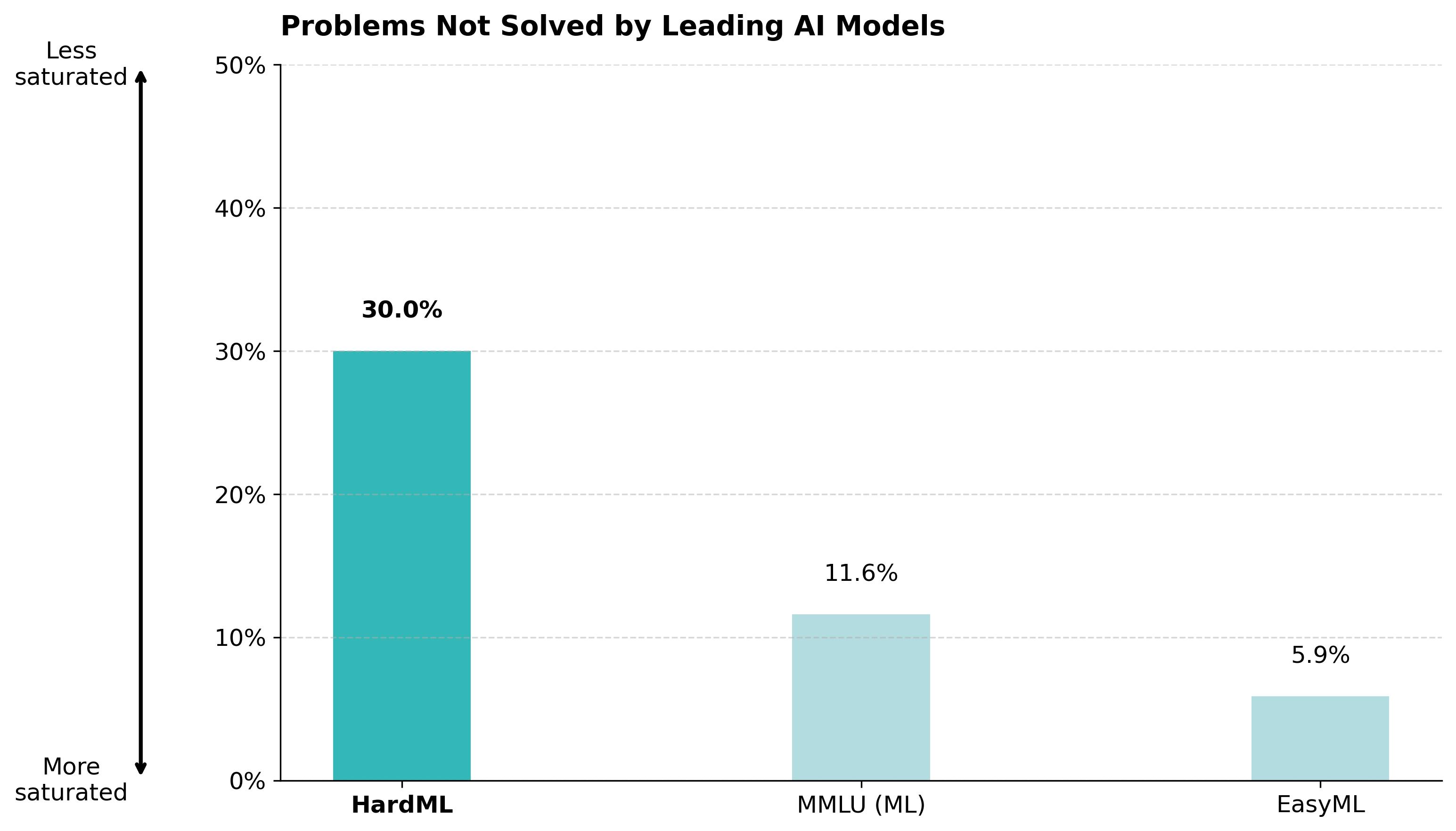}
\caption{Comparison of error rate across 3 DS\&ML benchmarks. While existing benchmarks are approaching saturation, HardML keeps an average level above saturation, in line with benchmarks from other fields like MathVista \cite{23} or AIME \cite{24} despite the multiple-choice nature}
\label{fig:unsolved_comparison}
\end{figure}

Our evaluation of state-of-the-art AI models (o1) \cite{20}, reveals a substantial performance gap compared to existing benchmarks [Figure~\ref{fig:unsolved_comparison}]. Specifically, these models exhibit an error rate of approximately 30\% on HardML, which is significantly higher than the error rate on the machine learning section of MMLU (MMLU-ML) \cite{5}. This disparity highlights the challenges that current AI models face in mastering the complexities of DS and ML, particularly in understanding nuanced concepts and applying them to non-trivial problems.

The initial motivation behind constructing this benchmark was to generate a comprehensive set of interview-preparation questions for individuals seeking positions in machine learning at leading technology companies (FAANG). However, the interesting results obtained during large language model evaluation, purely out of curiosity, led to the development of this paper. Given the relative scarcity of specialized benchmarks in these fields compared to others like mathematics and physics, we believe HardML fills an important gap and provides a foundation for future research and development.

%In parallel, recent work has begun to explore the capabilities of AI agents in machine learning engineering tasks. Notably, MLE-bench [14] evaluates AI agents on practical ML engineering skills using a collection of 75 Kaggle competitions. %This benchmark focuses on assessing agents' abilities to perform tasks such as data preparation, model training, and experimental analysis. While MLE-bench provides valuable insights into the practical engineering competencies of AI %agents, HardML complements this by focusing on theoretical knowledge and reasoning in DS and ML. Together, these benchmarks offer a more comprehensive evaluation of AI capabilities across both theoretical understanding and %practical application.

\section{Data collection}

The data collection involved a multi-step process spanned over 6 months. As mentioned in the last paragraph of the introduction, the initial purpose of this project was to form a set of question-answer for ML interview assessment for entrance of the top tech companies. These are to be used on the platform neuraprep.com, which is a website similar to leetcode.com for interview preparation. Therefore, the dataset construction wasn't biased towards building problems that the LLMs wouldn't be able to solve, they were fully intended for human use.

\subsection{The collection pipeline}

The collection pipeline for the development of HardML and EasyML involved a meticulous 4-step process:

\begin{enumerate}
    \item \textbf{Raw data collection and scraping.} We have sourced approximately 400 questions from public platforms such as Glassdoor, Blind, Quora, Stack Exchanges, YouTube, as well as from papers and books —including those by Bishop \cite{11}—and from our own writings and public blogs (\href{https://ptidor.com/single5.html}{ptidor.com}), among many other reputable sources. As such, we specifically dedicated time in collected ideas from modern sources - very recent interviews on the topic of the latest development in Natural Language Understanding (NLU) or Computer Vision (CV) and collecting ideas from recently published papers (from 2024 and 2023).

Importantly, this sourcing was not limited to simply gathering existing interview questions. Many questions were thoughtfully devised by us, inspired by theoretical concepts presented in books, papers, and online resources. This approach ensured that we had a reasonable amount of questions that were both original and rooted in fundamental principles of data science and machine learning.
    \item \textbf{Devising golden solutions and refinement.} In this phase, we crafted definitive "golden" answers for each question, providing clear and accurate solutions. Given that many sourced questions lacked reliable and complete answers, this was a demanding and iterative process that occupied the majority of the six-month development period.

Each question was paired with a golden answer and a list of core ideas—the essential elements required for a respondent to achieve a perfect score. During this stage, we also engaged in refining the questions, which included paraphrasing and restructuring to enhance clarity and coherence. However, to preserve the authenticity of real-world interview scenarios (recall that this was the purpose of this project at that time), not all questions were extensively modified; in some cases, we made only slight adjustments while maintaining the original intent. Upon completion, this raw dataset amounted to an extensive collection exceeding \textbf{150} pages (in google docs) of written material.

    \item \textbf{Adaptation to Multiple-Choice Format.} This is penultimate step of the process, and it involves transforming the refined dataset of questions, golden answers and core ideas into machine parsable/verifiable input and output. We chose to go with the MMLU (ML) framework of  multiple-choice question format, with a small change: at least one answer is correct, instead of exactly one that is correct, hence increasing the difficulty. This required converting the answers and core ideas into a structure where at least one option was correct. This process was non-trivial, as not all questions could be adapted without compromising their essence and level of difficulty. As a result, only about half of the initial questions and answers were successfully transitioned into the multiple-choice format, ensuring that the benchmark remained challenging and faithful to its original purpose.

    \item \textbf{Quality control and data contamination prevention.} The is the final step of the process,  focusing on rigorous quality assurance and final checks. We meticulously reviewed each multiple-choice question and corresponding answer to ensure accuracy, clarity, and alignment with the benchmark's objectives. This phase involved collaboration with beta testers—colleagues and peers—who interacted with the questions through the user interface (UI) of our platform (that the project was initially intended for), neuraprep.com. While no major errors were identified, several ambiguous cases were detected and rectified during this stage, enhancing the robustness and reliability of the final benchmark.

Finally, we conducted a contamination check by evaluating the similarity of our content against existing internet sources to detect potential plagiarism. If any high similarities were identified, the questions and answers were carefully adjusted to ensure originality. Note that this step was applied only to \textbf{HardML,} as its purpose is to rigorously assess human ML experts. In contrast, EasyML is intended to test the foundational knowledge and basic reasoning abilities of human entry-level professionals in DS and ML (and potentially smaller language models), and therefore, strict rigor was a secondary consideration.
\end{enumerate}

\subsection{Question difficulty}

The difficulty assignment to each question (between Easy, Medium and Hard) was done by us, as a measure of how difficult a question would appear in our eyes. The author of this paper is a former Lead Machine Learning Engineer with an MSc in AI from The University of Edinburgh with about 5 years of industry experience in machine learning. His research contributions have gathered over 80 citations  and his skill set encompasses a broad range of technologies and methodologies, including Python, PyTorch, AWS, GCP, MLOps, distributed computing, and quantitative finance. Most importantly, the author interviewed over 100 candidates throughout his career, driven by a deep passion for interview assessment and a commitment to excellence.

While we acknowledge and don't refute that the difficulty assignments may exhibit slight bias—given that they were determined by a single individual—we have strived to represent the difficulties as accurately as possible. This is substantiated by the results of our benchmark evaluations: HardML yields a significantly higher error rate than MMLU, indicating a higher level of difficulty, whereas EasyML achieves a notably lower error rate. These outcomes corroborate our assessments of the relative difficulties of the benchmarks. HardML comprises only "hard" questions (according to the categorization system explained above) whereas MMLU-ML --though lacking an official difficulty rating--appears to consist predominantly of "easy" and "medium" questions (according to the same categorization system).

\section{Dataset composition}

The HardML benchmark covers a broad spectrum of contemporary Data Science and Machine Learning spanning from basic data handling methods and classical machine learning to the frontier of Deep Learning and Natural Language Understanding with state-of-the-art language models and modern training pipelines utilizing tens of thousand of devices.

 \subsection{Dataset Statistics}

The distribution over categories is shows in [Table~\ref{tab:msc_distribution}]. A comprehensive coverage of topics is essential for effectively evaluating AI systems. Accordingly, the majority of the questions in our benchmark focus on Deep Learning, Natural Language Understanding (NLU), and Computer Vision (CV). This emphasis is intentional and natural, as these fields encompass the most novel approaches and present some of the most challenging questions in contemporary AI research. This distribution is in line with other prominent benchmarks' distributions like FrontierMath \cite{15}.

\begin{table}[h!]
\centering
\begin{tabular}{|p{6cm}|c|p{6cm}|c|}
\hline
\textbf{Category} & \textbf{Percentage}  \\ \hline
Deep Learning & 33\%  \\ \hline
Classical Machine Learning & 21\% \\ \hline
Natural Language Understanding & 15\%  \\ \hline
Data Engineering & 11\% \\ \hline
Computer Vision & 11\%  \\ \hline
Statistics \& Statistical Modeling & 9\% \\ \hline
\end{tabular}
\caption{Percentage distribution of DS and ML sub-fields in the HardML dataset, representing the proportion of each classification relative to the total amount.}
\label{tab:msc_distribution}
\end{table}

 \subsection{Comparison to related benchmarks}

HardML differs from the baseline MMLU benchmark in both size—being slightly smaller—and format: each question in HardML may have more than one correct answer. This multi-answer format also sets it apart from MLE-bench, which focuses on coding tasks with a definite answer rather than multiple-choice questions. A detailed comparison of the various datasets used in the research space for LLM evaluation in machine learning is presented in Table~\ref{tab:datasets_distribution}.

\setlength{\textfloatsep}{10pt} % Adjust spacing between text and floats (reduce it)

\begin{table}[h!]
\centering
\begin{tabular}{|m{5cm}|c|m{4cm}|c|}
\hline
\textbf{Dataset} & \textbf{Size}  & \textbf{Type}  \\ \hline
HardML (this paper) & 100  & multiple-choice \\ \hline
EasyML (this paper) & 85  & multiple-choice \\ \hline
MMLU [ML subset, test] & 112  & multiple-choice \\ \hline
MLE-bench (OpenAI) & 75 & coding \\ \hline
\end{tabular}
\caption{Comparison between datasets available in the research space for LLM evaluation in the field of DS and ML.}
\label{tab:datasets_distribution}
\end{table}

\vspace{-10pt} % Reduce space after the table

\subsection{Sample questions from HardML}
% Continue with your text here

In order to accurately provide an intuition of the level of difficulty and form of the questions from HardML, we display below a few examples.

% Problem 1
\begin{tcolorbox}[myboxstyle, title=Sample problem 1, fontupper=\small]
\textbf{Question:} You want to train an LLM that can solve challenging math problems properly. To do that, you employ a team of mathematicians to devise problems and solutions for training data. Unfortunately, you require a lot of training data, naturally, and hence you have to employ thousands of people to generate problems and solutions for your LLM. You need some form of quality control to understand if the mathematicians keep an overall good quality and that your LLM won't be trained on corrupted data. You can assume you have 1000 people devising (problem, solution) tasks, one person submits one task. Each task is rated from 5 choices, from 1/5 (lowest) to 5/5 (highest): 1/5,2/5,3/5,4/5,5/5. You want these people to produce, on average, a quality of work of at least 4/5=0.8 and to be 95\% sure that is the case. You cannot check all 1000 and compute the average yourself because that would defeat the purpose of employing these people in the first place, so then what's the minimum number N of random tasks you would need to check? For this exercise, you can assume that the task grades follow a normal distribution and the variance of the overall quality is known and it's the maximum it can be (each grade happens at least once), given the range 1/5-5/5. Make sure to normalize the grades in [0.2,1]

\begin{enumerate}[label=\textbf{\Alph*})]
    \item 4
    \item 6
    \item 7
    \item 8
\end{enumerate}

\textbf{Answer:} B
\end{tcolorbox}

% Problem 2
\begin{tcolorbox}[myboxstyle, title=Sample problem 2, fontupper=\small]
\textbf{Question:} You measure Model FLOPs Utilisation (MFU) by counting all floating point operations in the entire training step—including overhead—and dividing by (time elapsed)×(theoretical hardware FLOPS). You now enable activation (gradient) checkpointing, which re-runs parts of the forward pass to save memory. Assuming you still count all FLOPs and include the extra recomputations in your total, what will happen to your measured MFU?

\begin{enumerate}[label=\textbf{\Alph*})]
    \item MFU will strictly increase, because you are performing additional FLOPs without proportionally more time.
    \item MFU will strictly decrease, because the added time from redoing the forward pass dominates.
    \item MFU will remain exactly the same, because both FLOPs and time scale in a fixed ratio.
    \item The effect on MFU is ambiguous; you are doing more FLOPs but also increasing the total step time, so the ratio could go up or down.
\end{enumerate}

\textbf{Answer:} D
\end{tcolorbox}

% Problem 3
\begin{tcolorbox}[myboxstyle, title=Sample problem 3, fontupper=\small]
\textbf{Question:} A T5 or FlanT5 model is considered one of the best encoder-decoder models out there (as of 2024). Why aren't these commonly used at scale to train large language models (LLMs) that compete with GPT-4? Select all that apply.

\begin{enumerate}[label=\textbf{\Alph*})]
    \item The architecture of FlanT5 makes it harder to scale.
    \item Decoder-only models allow for simpler partitioning strategies, such as splitting along head dimensions, resulting in more balanced compute, memory, and network costs.
    \item T5 is like a sequence of blocks but with more edges representing more complicated data dependencies during compute.
    \item The communication between encoder and decoder in encoder-decoder models complicates network architecture and scaling strategies.
\end{enumerate}

\textbf{Answer:} A, B, C, D
\end{tcolorbox}

% Problem 4
\begin{tcolorbox}[myboxstyle, title=Sample problem 4, fontupper=\small]
\textbf{Question:} What is the difference between L2 regularization and weight decay in the context of neural networks, and under which conditions can they be considered equivalent?

\begin{enumerate}[label=\textbf{\Alph*})]
    \item L2 regularization adds a penalty to the loss function proportional to the sum of squared weights, while weight decay multiplies the weights by a factor slightly less than 1 after each update.
    \item L2 regularization and weight decay are always equivalent, regardless of the optimizer used.
    \item L2 regularization and weight decay are equivalent only when using stochastic gradient descent (SGD) as the optimizer.
    \item Using optimizers like Adam or RMSprop breaks the equivalence between L2 regularization and weight decay.
\end{enumerate}

\textbf{Answer:} A, C, D
\end{tcolorbox}

% Problem 5
\begin{tcolorbox}[myboxstyle, title=Sample problem 5, fontupper=\small]
\textbf{Question:} The backpropagated gradient through a tanh non-linearity is always smaller or equal in magnitude than the upstream gradient.

\begin{enumerate}[label=\textbf{\Alph*})]
    \item True.
    \item Depends on the sign of the inputs.
    \item False
    \item True only if all the input units are in (-1,1)
\end{enumerate}

\textbf{Answer:} A
\end{tcolorbox}

% Problem 6
\begin{tcolorbox}[myboxstyle, title=Sample problem 6, fontupper=\small]
\textbf{Question:} Where is the temperature applied in the model architecture of Chat GPT-3 or 4?

\begin{enumerate}[label=\textbf{\Alph*})]
    \item At the input level.
    \item After the softmax layer.
    \item Right before the softmax layer.
    \item At beam-search level when we select the output token based on probability.
\end{enumerate}

\textbf{Answer:} C
\end{tcolorbox}

We intentionally designed the benchmark to assess fairly up-to-date advancements in AI, as exemplified by questions \textbf{2, 3, and 6}. Additionally, we included both reasoning-intensive questions—such as question \textbf{1}, which requires code implementation or meticulous hand calculations, and question \textbf{5}, which tests comprehension through a comparison between the hyperbolic tangent function (tanh) and its derivative—as well as knowledge-intensive questions like question \textbf{4}, which addresses a subtle nuance in the mathematical formula for weight decay and the formula of popular optimizers.

\section{Results}

\subsection{Accuracy on HardML}

\begin{figure}[h]
\centering
\includegraphics[width=1\textwidth]{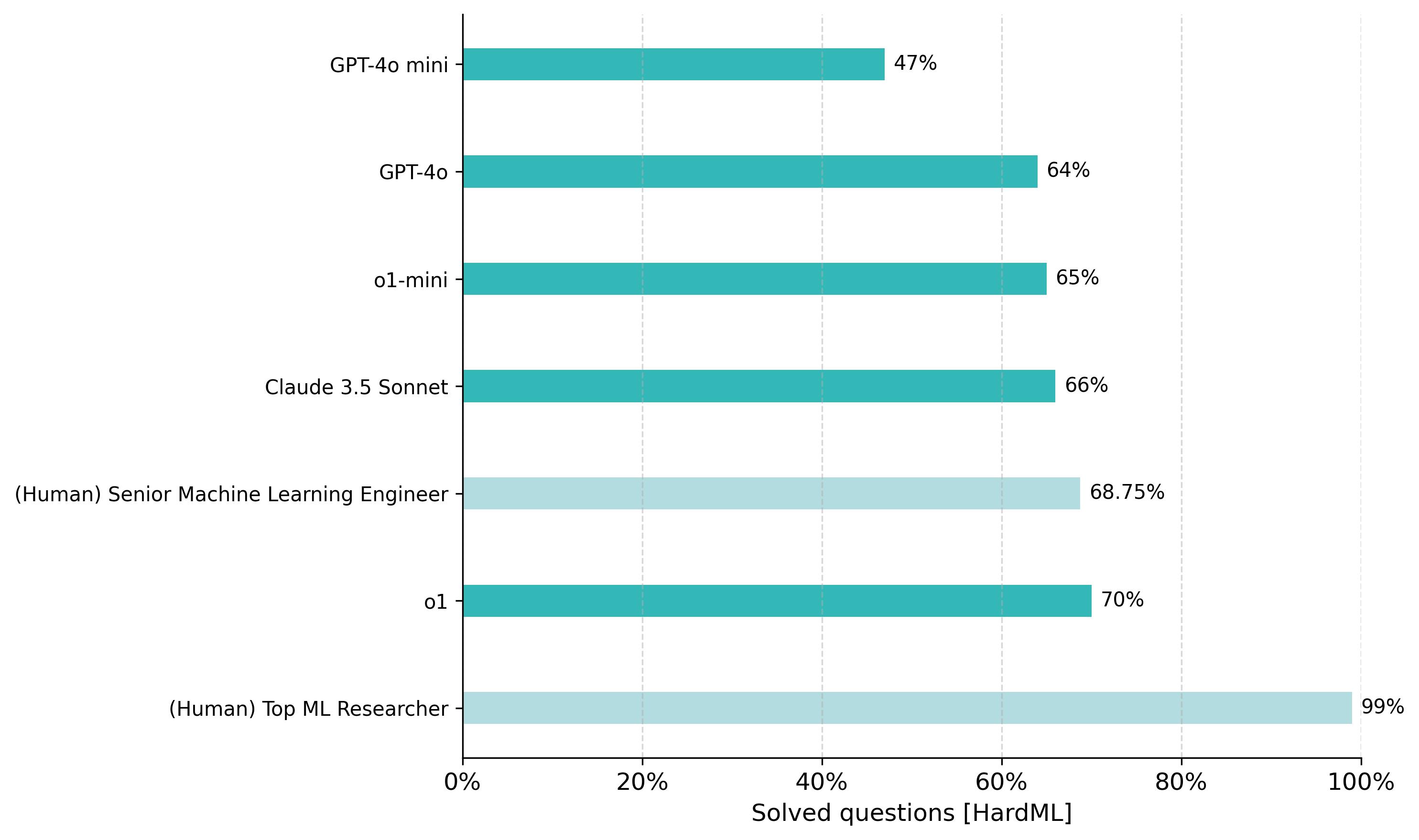}
\caption{Solved questions in HardML}
\label{fig:hardml_comparison}
\end{figure}

We evaluated 5 leading language models and 1 leading smaller language model (gpt-4o-mini) on our HardML dataset. Due to limited resources and ease of use, we decided to stick only with models from OpenAI and Anthropic, we believe these are enough to convey a good assessment. For instance, o1 is in top 5 in Chatbot Arena LLM Leaderboard from lmarena.ai. The results are present in [Figure~\ref{fig:hardml_comparison}]. We used the same prompt and batch size for these experiments.

Based on a single evaluation of the full benchmark, we found that most models solve aout \textbf{65\%} of the questions with the top performing model (o1) being able to solve \textbf{70\%}. This is in line with other benchmarks from other fields. For example, in math, current benchmarks that are considered 'hard' like Omni-Math, MathVista and Aime have around \textbf{60\%}, \textbf{70\%} and around \textbf{70\%} respectively in accuracy against o1. Interestingly, if we allow a "soft" figure for solved questions (giving partial credit when the model's answer is a subset of the correct answer), then the performance goes up by \textbf{5} percentage points (to \textbf{75.08\%} for o1), not a drastic change.

The models are very close together in performance, the precise ordering of model performance should be interpreted with significant caution as multiple runs could switch a few places around. We ran 2 times to make sure the order at the top is consistent, in both, o1 demonstrated the strongest performance.

An interesting observation is that even when the model arrives at a correct result, the underlying reasoning may not be accurate. We made the models output a reasoning field in the output json to observe this behaviour. For example, GPT-4o sometimes selects numerical answers because they are intuitively closer to an expected magnitude (like choosing 7 over 8 because it is smaller), rather than deriving them through rigorous proof. This illustrates a natural limitation of the multiple-choice format—scores can be artificially inflated due to luck or educated guesses that do not reflect true understanding.

\begin{figure}[h]
\centering
\includegraphics[width=1\textwidth]{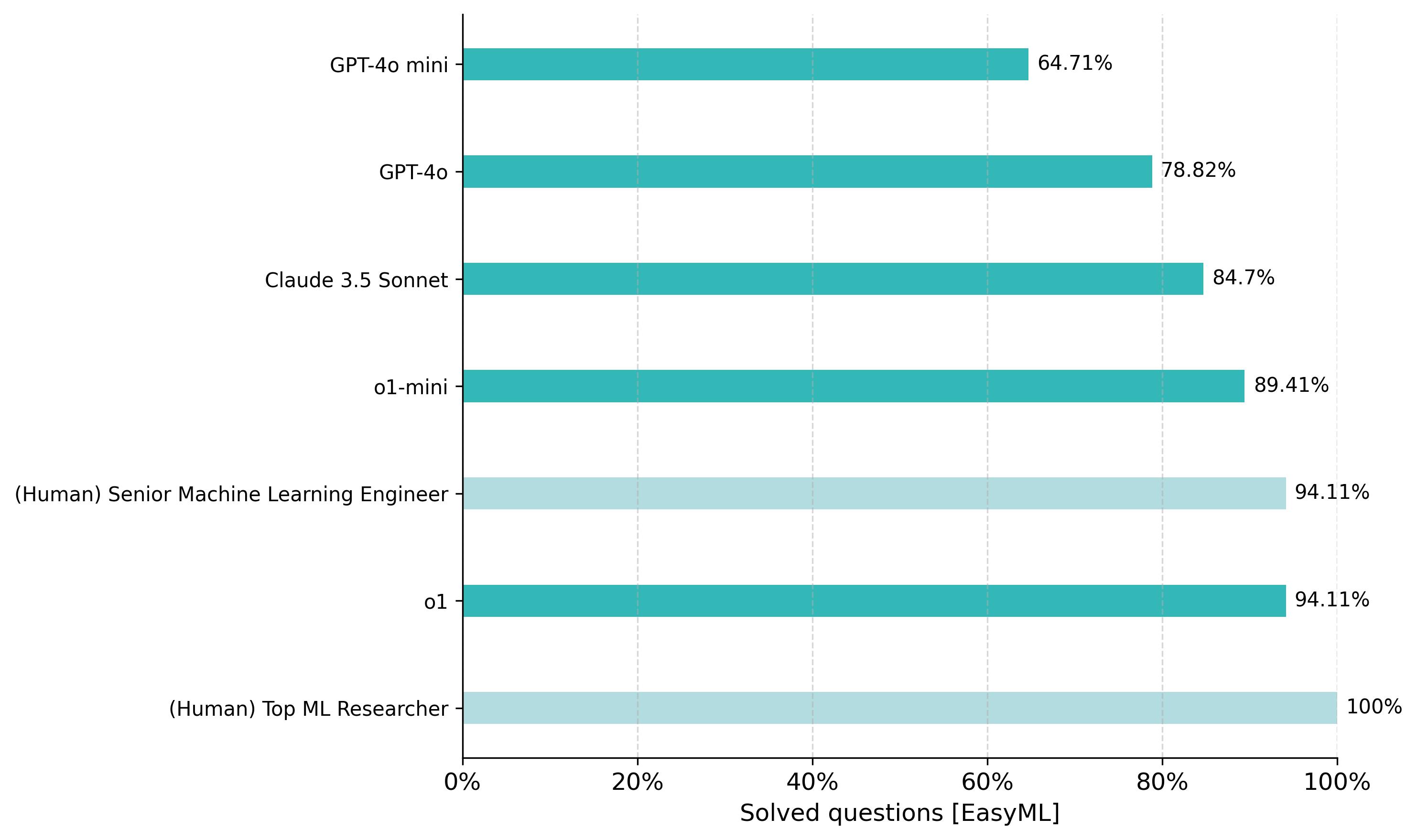}
\caption{Solved questions in EasyML}
\label{fig:easyml_comparison}
\end{figure}

\subsection{Human performance on HardML}

Examining human performance is particularly insightful, given that the initial intention of this project was to assess candidates during interviews or to filter applicants competing for positions in major technology companies. The results displayed in Figure~\ref{fig:hardml_comparison} include human scores for reference. Below, we explain how these human scores were calculated:

\begin{enumerate}[label=\textbf{\Alph*})]
    \item The first metric (Senior Machine Learning Engineer) was obtained during the beta testing phase of data collection (the final step). We invited actual senior machine learning engineers—friends of the author (see acknowledgements)—to participate in several quizzes consisting of 7 to 8 questions each, sampled, at random, from HardML. Once sampled, the same quizzes were used for each person, we managed to assess 5-6 quizzes per person. After aggregating the results, we found that an individual scored, on average, approximately 5.5 correct answers out of 8 (which translates into 68.75\%). Although it is not reflective of the performance on all the questions from HardML (only on a subset), we believe this figure is relevant to be shown. Hence, this performance is reflected in Figure~\ref{fig:hardml_comparison} and Figure~\ref{fig:easyml_comparison}. The participants expressed admiration for the benchmark, noting that the questions required significant thought and were highly challenging.
    \item The second number (Top ML Researcher) is purely the author's opinion. Even though we did not have access to a globally recognized top machine learning researcher, we posit that this benchmark would not represent a significant challenge for individuals actively engaged in cutting-edge ML research and who have been at the forefront of the field for the past 20 years.
\end{enumerate}

\subsection{Accuracy on MMLU and EasyML}

Below, we have the equivalent diagram (Figure~\ref{fig:mmlu_comparison}) for the 112 questions present in the testset of MMLU (ML subset) and our proposed EasyML. Observe how o1 is still the top performer, but the scores are significantly higher compared to HardML. Note that, we have not displayed human assessment figures on the MMLU benchmark as this experiment wasn't conducted.

\begin{figure}[h]
\centering
\includegraphics[width=1\textwidth]{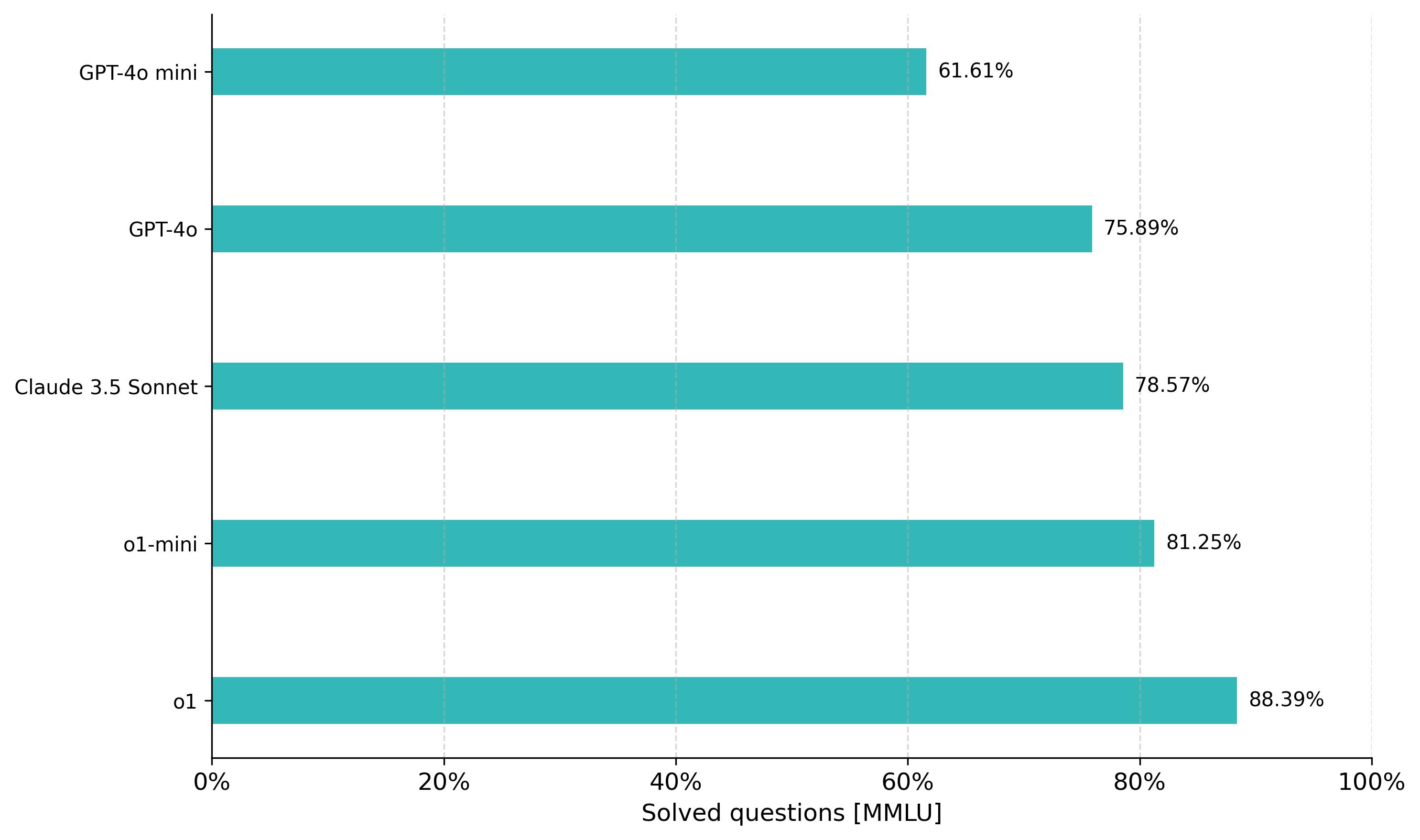}
\caption{Solved questions in MMLU [ML]}
\label{fig:mmlu_comparison}
\end{figure}

\section{Related work}

The evaluation of large language models (LLMs) has been extensively explored across various domains, leading to the development of numerous benchmarks that assess different aspects of AI capabilities. In this section, we review the most relevant benchmarks and studies related to our work, focusing on those that evaluate LLMs in the context of machine learning and data science and briefly mentioning a few impressive pieces of work on other fields from STEM.

\subsection{Multitask Language Understanding Benchmarks}

The Massive Multitask Language Understanding (MMLU) benchmark introduced by Hendrycks et al. \cite{5} has been a significant step toward assessing the broad academic and professional knowledge of LLMs. MMLU covers 57 subjects across STEM, humanities, social sciences, and more, including a subset dedicated to machine learning with 112 multiple-choice questions. While MMLU has provided valuable insights into the capabilities of models like GPT-3, state-of-the-art models have begun to approach saturation on several subjects, including ML. This near-ceiling performance limits the benchmark's effectiveness in distinguishing the advanced capabilities of newer models. Moreover, the ML subset, due to its relatively small size and scope, may not fully capture the depth and complexity required to evaluate nuanced understanding in ML and DS.

\subsection{Benchmarks for Advanced Reasoning}

To address the limitations of existing benchmarks in measuring advanced reasoning, FrontierMath \cite{15} was introduced as a benchmark comprising exceptionally challenging and original mathematical problems. These problems span major branches of modern mathematics and are designed to require significant effort from expert mathematicians—often multiple hours or days—to solve. FrontierMath effectively minimizes data contamination by using unpublished problems and employs automated verification for reliable evaluation. Remarkably, current AI models solve under 2\% of the problems, highlighting a substantial gap between AI capabilities and human expertise in advanced mathematics. This benchmark underscores the importance of creating future-proof evaluations that remain challenging despite rapid advancements in AI. This paper is the inspiration for HardML, we were impressed by it and wanted to replicate some of the work.

\subsection{Practical Machine Learning Engineering Benchmarks}

In parallel, MLE-bench was proposed by Chan et al \cite{16} as a benchmark to evaluate AI agents' performance in machine learning engineering tasks. MLE-bench curates 75 ML engineering-related competitions from Kaggle, encompassing tasks that require practical skills such as data preprocessing, model training, and experimental analysis. By establishing human baselines based on Kaggle's publicly available leaderboards, MLE-bench provides a real-world context for assessing AI agents in practical engineering scenarios. The benchmark evaluates AI setups like OpenAI's o1-preview with AIDE scaffolding, noting that the best-performing agent achieves a bronze medal level in approximately 17\% of competitions.

\subsection{Automated Answering and Generation of ML Exams}

In the realm of educational assessments, other researchers explored the automatic answering and generation of machine learning final exam questions in their work titled "From Human Days to Machine Seconds: Automatically Answering and Generating Machine Learning Final Exams." \cite{25} They demonstrated that large language models could pass ML final exams at a human level and generate new exam questions rapidly. Their study focused on the differences between final exams and problem sets, noting that final exams typically have longer, multi-part questions that span a broader set of topics and require more complex reasoning. Notably, in this paper, multiple-choice questions were generated and tested, making it a valuable related benchmark that is, in our opinion, underexplored.

\subsection{Comparison to Our Work}

Our proposed HardML benchmark fills an important gap in existing evaluations by providing a rigorous, modern, and challenging testbed specifically tailored to data science and machine learning. Unlike MMLU's ML subset, HardML offers a more difficult, more diverse and more up-to-date set of questions that delve deeper into advanced topics. In contrast to MLE-bench, which assesses practical engineering skills through coding tasks, HardML focuses on theoretical understanding and the ability to reason about complex concepts.

By emphasizing originality and minimizing data contamination, similar to FrontierMath, we ensure that HardML remains a relevant and challenging benchmark for current and future AI models. Additionally, by including EasyML as a complementary benchmark for evaluating smaller language models, we address the need for scalable evaluations across different model sizes and capabilities. It is challenging to ascertain the long-term applicability of HardML; however, we anticipate that it will remain relevant at the cutting edge of model evaluation for at least one year.

\section{Limitations}

Even though HardML currently demonstrates reasonable resistance to saturation, we do not believe this resilience will persist for much longer. Models like o3 \cite{26} have already shown improvements over previous frontier models such as o1, and the pace of advancement in AI systems is exceedingly rapid. One of the significant limitations of HardML is its multiple-choice format, which allows for "guesses" or "educated guesses" that can artificially inflate scores—a limitation that has been critically examined in FrontierMath. In benchmarks like MMLU [ML], where only one answer is correct per question, a random guess has a  $\frac{1}{4}$ chance of being correct. In comparison, in a multiple-choice format where more than one answer can be correct, a random guess has a probability as high as $\frac{1}{15}$ . These probabilities are still substantial, potentially diminishing the benchmark's ability to effectively discriminate between true understanding and chance performance.

Therefore, it is essential to develop benchmarks with automatic evaluations that require machine-verifiable outputs, such as numerical or boolean answers. This approach reduces the likelihood of inflated scores due to guessing. Benchmarks like MLE-bench, which necessitate code implementation or involve advanced mathematical reasoning to arrive at the correct solution—while still being related to data science and machine learning—are exemplary in this regard.

Constructing challenging multiple-choice questions is particularly difficult because adept humans or advanced AI models can employ elimination strategies to identify the correct answers. This means that even if the correct answers are difficult to determine, the benchmark's effectiveness can collapse if the incorrect options are not \textbf{equally challenging} to dismiss. Consequently, every answer choice must be nuanced and not obviously incorrect. Achieving this level of subtlety in question design is exceptionally demanding and was a primary reason why the development of this benchmark required such a significant investment of time. Crafting answers that appear plausible yet are subtly incorrect is a skill in itself.

\section{Acknowledgements}

Special thanks to: Robin Kahlow (Senior ML Engineer at RunwayML), Geo Badita (Senior Software Engineer at Meta), and Chady Dimachkie (Head of ML at Abwab.ai and former Deep Learning Engineer at Nvidia) for taking the challenge and attempting HardML very thoroughly, as well as providing invaluable feedback. Their expertise and rigorous assessments have been instrumental in refining the dataset and validating its efficacy. Special thanks to Paul Chelarescu for his invaluable assistance in curating and organising the raw database in the first step of data collection, which served as the foundation for this work.

\section{Conclusion}

With this paper, we instigate to further research in the area of LLM benchmarking for cutting edge Data Science and Machine Learning.  The dataset of HardML is present in an interactive environment on \href{https://neuraprep.com}{neuraprep.com} and can also be obtained in clean json format for experiment replication or further research \href{https://drive.google.com/file/d/1hTiM1wu_enxAVVVtgwR8Fzaa3Ff2zEFI/view?usp=sharing}{here}. Our work contributes to the ongoing efforts to develop benchmarks that can effectively measure and distinguish the advanced capabilities of AI models in rapidly evolving fields.

\end{document}